\documentclass[10pt,twocolumn,letterpaper]{article}

\usepackage{cvpr}
\usepackage{times}
\usepackage{epsfig}
\usepackage{graphicx}
\usepackage{amsmath}
\usepackage{mathtools}
\usepackage{threeparttable}
\usepackage{amssymb}
\usepackage{multirow}
\usepackage{algorithm}
\usepackage{balance}
\usepackage[noend]{algpseudocode}

\makeatletter
\def\algbackskip{\hskip-\ALG@thistlm}
\makeatother

\usepackage[pagebackref=true,breaklinks=true,letterpaper=true,colorlinks,bookmarks=false]{hyperref}

\cvprfinalcopy 

\def\cvprPaperID{5} 

\ifcvprfinal\fi

\begin{document}

\title{Enhancing Cross-task Transferability of Adversarial Examples with Dispersion Reduction}

\author{
Yunhan Jia\thanks{Equal contribution}, Yantao Lu\textsuperscript{*}$^\dagger$, Senem Velipasalar$^\dagger$, Zhenyu Zhong, Tao Wei\\
Baidu X-Lab, $^\dagger$Syracuse University
\\
\{jiayunhan, yantaolu, edwardzhong, lenx\}@baidu.com, svelipas@syr.edu
} 

\maketitle

\begin{abstract}
Neural networks are known to be vulnerable to carefully crafted adversarial examples, and these malicious samples often transfer, i.e., they maintain their effectiveness even against other models. With great efforts delved into the transferability of adversarial examples, surprisingly, less attention has been paid to its impact on real-world deep learning deployment. 

In this paper, we investigate the transferability of adversarial examples across a wide range of real-world computer vision tasks, including image classification, explicit content detection, optical character recognition (OCR), and object detection. It represents the  cybercriminal's situation where an ensemble of different detection mechanisms need to be evaded all at once.

We propose practical attack that overcomes existing attacks' limitation of requiring task-specific loss functions by targeting on the ``dispersion'' of internal feature map. We report evaluation on four different computer vision tasks provided by Google Cloud Vision APIs to show how our approach outperforms existing attacks by degrading performance of multiple CV tasks by a large margin with only modest perturbations ($l_\infty\leq16$).

\end{abstract}
\section{Introduction}
Recent research in adversarial learning has brought the weaknesses of deep neural networks (DNNs) to the spotlights of security and machine learning studies. Given a deep learning model, it is easy to generate adversarial examples (AEs), which are close to the original but are misclassified by the model~\cite{carlini2017towards, szegedy2013intriguing}. More importantly, their effectiveness sometimes \emph{transfer}, which may severely hinder DNN based applications especially in security critical scenarios~\cite{liu2016delving, dong2018boosting, xie2018improving}. While such vulnerabilities are alarming, little attention has been paid on the realistic threat model of commercial or proprietary vision-based detection systems against real-world cybercriminals, which turn out to be quite different from those intensively studied by aforementioned research.

\begin{figure}
\centering
\vspace{0.5cm}
\includegraphics[width=\columnwidth]{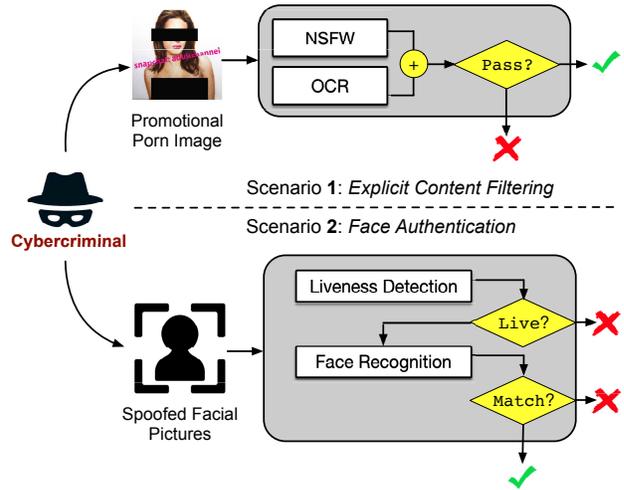}
\caption{Real-world computer vision systems deployed in safety- and security-critical scenarios usually employ an ensemble of detection mechanisms that are opaque to attackers. Cybercriminals are required to generate adversarial examples that transfer across tasks to maximize their chances of evading the entire detection systems.}
\label{fig:threat}
\vspace{-0.4cm}
\end{figure}

\textbf{Deployment.} Computer vision (CV) based detection mechanisms have been extensively deployed in security-critical applications such as content censorship and authentication with facial biometrics, and readily available services are provided by cloud giants through APIs (e.g., Google Cloud Vision~\cite{google}, Amazon Rekognition~\cite{amazon}). The detection systems have long been targeted by evasive attacks from cybercriminals, and it has resulted in an arm race between new attacks and more advanced defenses.

\textbf{Ensemble of different detection mechanisms.} To overcome the weakness of deep learning in individual domain, real-world CV systems tend to employ an ensemble of different detection mechanisms to prevent evasions. As shown in Fig.~\ref{fig:threat}, underground businesses embed promotional contents such as URLs into porn images with sexual content for illicit online advertising or phishing. A detection system combines Optical Character Recognition (OCR) and image-based explicit content detection can thus drop posted images containing either suspicious URLs or sexual content to mitigate evasion attacks. Similarly, a face recognition model that is known to be fragile~\cite{sharif2016accessorize} is usually protected by a liveness detector to defeat spoofed digital images when deployed for authentications. Such ensemble mechanisms are widely adopted in real-world CV deployment.

To evade detections with uncertain mechanisms, attackers turn to generate adversarial examples that transfer across CV tasks. Many adversarial techniques on enhancing transferability have been proposed~\cite{zhou2018transferable, xie2018improving, liu2016delving, dong2018boosting}. However, most of them are designed for image classification tasks, and rely on task-specific loss function (e.g., cross-entropy loss), which limits their effectiveness when transferred to other CV tasks.

\begin{figure}
\centering
\includegraphics[width=\columnwidth]{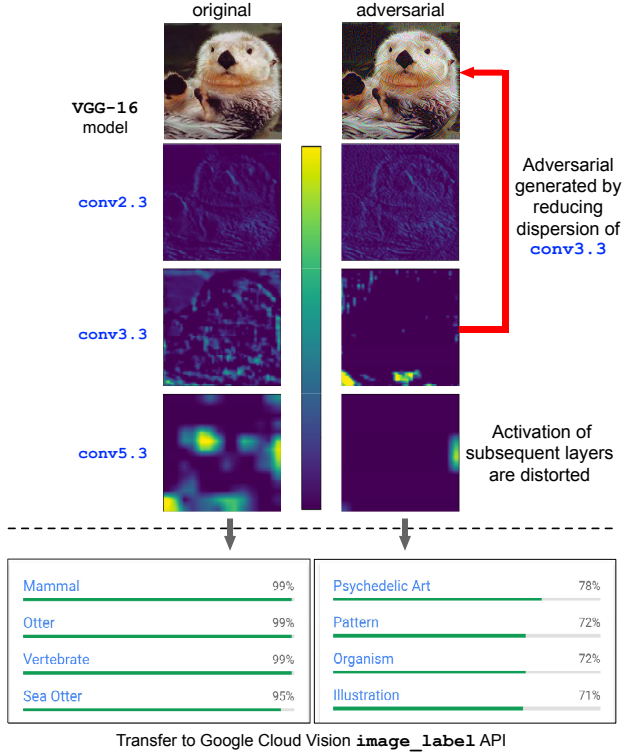}
\caption{DR attack targets on the dispersion of feature map at specific layer of feature extractors. The adversarial example generated by minimizing dispersion at \texttt{conv3.3} of VGG-16 model also distorts feature space of subsequent layers (e.g., \texttt{conv5.3}), and its effectiveness transfers to commercially deployed GCV APIs.}
\label{fig:illustration}
\end{figure}

In this paper, we propose a simple yet effective approach to generate adversarial examples that transfer across a broad class of CV tasks, including classification, object detection, explicit content detection and OCR. Our approach called \emph{Dispersion Reduction} (\textbf{DR}) as shown in Fig.~\ref{fig:illustration}, is inspired by the impact of ``contrast'' on an image's perceptibility. As lowering the contrast of an image would make the objects indistinguishable, we presume that reducing the ``contrast'' of internal feature map would also degrade the recognizability of the subjects in the image, and thus could evade CV-based detections. We use \emph{dispersion} as a measure of ``contrast'' in feature space, which describes how scattered a set of data is. We empirically validate the impact of dispersion on model predictions, and find that reducing the dispersion of internal feature map would largely affect the activation of subsequent layers. Based on another observation that lower layers detect simple features~\cite{lee2009convolutional}, we hypothesis that the low level features extracted by early convolution layers share many similarities across CV models. Thus the distortions caused by dispersion reduction in feature space, are ideally suited to fool any CV models, whether designed for classification, object detection, OCR, or other vision tasks. 

We evaluate our proposed DR attack on both popular open source models and commercially deployed detection models. The results on four Google Cloud Vision APIs: classification, object detection, SafeSearch, and OCR (see \S\ref{sec:experiments}) show that our attack causes larger drops on the model performance than state-of-the-art attacks ( MI-FGSM~\cite{dong2018boosting} and DIM~\cite{xie2018improving}) by a big margin of 11\% on average across different tasks. We hope that our finding to raise alarms for real-world CV deployment in security-critical applications, and our attacks to be used as benchmarks to evaluate the robustness of DNN-based detection mechanisms. Code is available at: \href{https://github.com/jiayunhan/dispersion\_reduction}{https://github.com/jiayunhan/dispersion\_reduction}.

\section{Background \& Related Work}

\subsection{Transferability of Adversarial Examples}

Since the seminal finding of Szegedy \etal~\cite{szegedy2013intriguing}, the transferability of adversarial examples between different models trained over same or disjoint datasets have been discovered. Followed by Goodfellow \etal~\cite{goodfellow2014explaining}, this phenomenon was attributed to the reason that adversarial perturbations is highly aligned with the weight vector of model. More recently, Papernot \etal~\cite{papernot2017practical} investigated attacks against black-box models by training substitute models. They also demonstrated attacks against machine learning services hosted by Amazon, and Google.

Our work differs from Papernot \etal~\cite{papernot2017practical} in two main aspects. First, the GCV APIs we attack in this work is not the same as the Cloud Prediction API~\cite{google_engine} (now the Google Cloud Machine Learning Service) attacked in Papernot \etal~\cite{papernot2017practical}. Both systems are black-box, but the Prediction API is intended to be trained by user's own data, while the GCV APIs are trained on Google's data and are provided "out-of-box". Second, we study transferability over black-box commercial models assuming no feedback on testing samples. Our proposed DR attack do not query the systems for constructing substitute model~\cite{papernot2017practical, papernot2016transferability} nor running score or decision based attacks~\cite{brendel2017decision, ilyas2018black, narodytska2016simple, su2019one}, and as Liu \etal~\cite{liu2016delving} demonstrated, it is more difficult to transfer adversarial examples to commercial models that are trained on large dataset, and are potentially ensemble. 

\subsection{Adversarial Attacks}
Several methods have been proposed recently to find AEs and improve transferability. A single-step attack, called fast gradient sign method (FGSM) was proposed by Goodfellow \etal~\cite{goodfellow2014explaining}. In a follow up work, Kurakin \etal~\cite{kurakin2016adversarial} proposed a multi-step attack, called iterative fast gradient sign method (I-FGSM) that iteratively searches the loss surface. Generally iterative attack achieves higher success rate than single-step attack in white-box setting, while performs worse when transfer to other models~\cite{xie2018improving}. 

Fueled by the NIPS 2017 adversary competition~\cite{kurakin2018adversarial}, several adversarial techniques that enhance transferability have been introduced, among them we given an overview of the most notable ones.

\textbf{MI-FGSM.} Momentum Iterative Fast Gradient Sign Method (MI-FGSM) proposed by Dong \etal~\cite{dong2018boosting} integrates momentum term into the attack process to stabilize update directions and escape poor local maxima. The update procedure is as follow:
\begin{equation}
\begin{aligned}
x'_{t+1} = x'_t + \alpha \cdot sign(g_{t+1}) \\
g_{t+1} = \mu \cdot g_{t} + \frac{\bigtriangledown_{x}J(x'_{t},y)}{ \parallel \bigtriangledown_{x}J(x'_{t},y) \parallel_1}
\end{aligned}
\label{eq:mifgsm}
\end{equation}
The strength of MI-FGSM can be controlled by the momentum and the number of iterations. 

\textbf{DIM.} Momentum Diverse Inputs Fast Gradient Sign Method (DIM) combines momentum and input diversity strategy to enhance transferability~\cite{xie2018improving}. DIM applies image transformation($T(\cdot)$) to the inputs with a probability $p$ at each iteration of iterative FGSM to alleviate the overfitting phenomenon. The updating procedure is similar to MI-FGSM, with the only replacement of Eq.\ref{eq:mifgsm} by:
\begin{equation}
\begin{aligned}
x'_{t+1} = Clip^\epsilon_x\{x'_t + \alpha \cdot sign(\bigtriangledown_x L(T(x'_{t+1};p),y^{true})\} \\
\end{aligned}
\label{eq:dim}
\end{equation}
where $T(x'_t,p)$ is a stochastic transformation function that performs input diversion on input with a probability of $p$.

The major difference between dispersion reduction (DR) with existing attacks is that DR doesn't require task-specific loss functions (e.g., cross-entropy used by the family of FGSM attacks). It targets on the numerical property of low level features that is task-independent, and presumably similar across CV models. Our evaluation in \S\ref{sec:experiments} demonstrate good transferability of adversarial examples generated by DR across real-world CV tasks.

\section{Methodology}

\begin{algorithm}
    \caption{Dispersion reduction attack}
    \textbf{Input:} A classifier $f$, original sample $x$, feature map at layer $k$; perturbation budget $\epsilon$ \\
    \textbf{Input:} Attack iterations $T$, learning rate $\ell$. \\
    \textbf{Output:} An adversarial example $x'$ with  ${\parallel x'-x\parallel}_\infty\leqslant\epsilon$
    \begin{algorithmic}[1]
    \Procedure{Despersion reduction}{}
    \State $x'_0 \gets x$
    \For{$t=0$ to $T-1$}
        \State Forward $x'_t$ and obtain feature map at layer $k$:
        \begin{equation}
        \mathcal{F}_k = f(x'_t)|_k
        \end{equation}
        \State Compute standard deviation of $\mathcal{F}_k$: $g(\mathcal{F}_k)$
        \State Compute its gradient $w.r.t$ the input: $\bigtriangledown_{x}g(\mathcal{F}_k)$
        \State Update $x'_t$ by applying Adam optimization:
        \begin{equation}
        x'_t = x'_t - Adam(\bigtriangledown_{x}g(\mathcal{F}_k), \ell)
        \end{equation}
        \State Project $x'_t$ to the vicinity of $x$:
         \begin{equation}
        x'_t = clip(x'_t, x-\epsilon, x+\epsilon)
        \end{equation}     
    \EndFor
    \State \textbf{return} $x'_t$ 
    \EndProcedure
    \end{algorithmic}
    \label{alg:dr}
\end{algorithm}

Existing attacks perturb input images along gradient directions $\bigtriangledown_x J$ that depend on the ground-truth label $y$ and the definition of the task-specific loss function $J$, which limits their cross-task transferability. We propose \emph{dispersion reduction} (DR) attack that formally define the problem of finding an AE as an optimization problem:

\begin{equation}
\begin{aligned}
&\min_{x‘} g(f(x',\theta)) \\
s.t.\   &{\parallel x'-x\parallel}_\infty\leqslant\epsilon \\
\end{aligned}
\label{eq:DR}
\end{equation}

where $f(\cdot)$ is a DNN classifier with output of intermediate feature map and $g(\cdot)$ calculates the dispersion. Our proposed DR attack in Algorithm~\ref{alg:dr} takes a multi-step approach that creates an adversarial example by iteratively reducing the dispersion of intermediate feature map at layer $k$. Dispersion describes the extent to which a distribution is stretched or squeezed, and there can be different measures of dispersion such as the variance, standard deviation, and gini coefficient~\cite{statistics}. In this work, we choose standard deviation as the dispersion metric and denote it as $g(\cdot)$ due to its simplicity.

Given a target feature map, DR applies Adam optimizer to iteratively perturb image $x'$ along the direction of reducing standard deviation, and projects it to the vicinity of $x$ by clipping at $x\pm\epsilon$. Denoting the feature map at layer $k$ as $\mathcal{F}_k = f(x'_t)|_k$, DR attack solves the following formula:

\begin{equation}
\begin{aligned}
x'_{t+1}&=x'_t-\bigtriangledown_{x'}{g(\mathcal{F}_k)} \\
&=x'_t-\frac{d g(t)}{dt}\cdot \frac{d f(x'_t)|_k}{dx'} \\
&=x'_t-\frac{t-\bar{t}}{\sqrt{N-1}\cdot\sqrt{\sum_{i}{(t_i-\bar{t})}^2}}\cdot\frac{d f(x'_t)|_k}{dx'}
\end{aligned}
\label{eq:DR_gradient}
\end{equation}

\begin{table*}[]
\centering
\begin{tabular}{|c|c|cccc|}
\hline
\multirow{2}{*}{} & \multirow{2}{*}{\begin{tabular}[c]{@{}c@{}}Target  Layer\end{tabular}} & \multicolumn{2}{c|}{Classification - acc.} & \multicolumn{2}{c|}{Detection - mAP(IoU=0.5)} \\ \cline{3-6}
 &  & \multicolumn{1}{c|}{Inception-v3} & \multicolumn{1}{c|}{DenseNet} & \multicolumn{1}{c|}{RetinaNet} & \multicolumn{1}{c|}{YOLOv3} \\ \hline \hline
\multicolumn{1}{|c|}{\multirow{3}{*}{VGG-16}} & conv1.2 (shallow)& 52.5\% & 29.3\% & 31.8 & 42.3 \\
\multicolumn{1}{|c|}{} & conv3.3 (mid) & \textbf{28.7}\% & \textbf{34.6}\% & \textbf{18.3} & \textbf{33.8} \\
\multicolumn{1}{|c|}{} & conv5.1 (deep)& 35.5\% & 44.8\% & 34.0 & 41.5 \\ \hline \hline
\multirow{3}{*}{Resnet-152} & conv1 (shallow)& 53.7\% & 63.1\% & 28.3 & 57.3 \\
 & conv3.8.3 (mid)& \textbf{25.8}\% & \textbf{34.7}\% & \textbf{29.5} & \textbf{41.6} \\
 & conv5.3.3 (deep)& 28.4\% & 41.5\% & 20.5 & 38.5 \\ \hline
\end{tabular}
\vspace{0.1cm}
\caption{\textbf{The performance of classification and object detection models (columns) when attacked by adversarial examples generated on VGG-16 and Resnet-152}. The profiling result suggests that AEs generated by targeting middle layers degrade performance of both classification and detection models by a larger margin.}
\label{tab:methodology}
\end{table*}

From Eq.\ref{eq:DR_gradient}, we state that given the targeted intermediate feature map, the optimized adversarial example $x'_t$ is achieved when all feature map elements $t_j\in t$ have the same value. Table~\ref{tab:result} compares the transferability of AEs generate on different layers (shallow to deep) of off-the-shelf feature extractors across different classification and object detection models. The result on 1000 randomly chosen samples from ImageNet validation set shows that targeting on middle layers, i.e. \texttt{conv3.3} of VGG-16 and \texttt{conv3.8.3} of Resnet-152 provides better transferability.

\section{Experiments}
\label{sec:experiments}
In this section, we compare DR with state-of-the-art adversarial techniques to enhance transferability on commercially deployed Google Cloud Vision (GCV) tasks: 
\begin{itemize}
    \item Image Label Detection (\textbf{\texttt{Labels}})~\footnote{https://cloud.google.com/vision/docs/detecting-labels} classifies image into broad sets of categories. 
    \item Object Detection (\textbf{\texttt{Objects}})~\footnote{https://cloud.google.com/vision/docs/detecting-text} detects multiple objects with their labels and bounding boxes in an image. 
    \item Image Texts Recognition (\textbf{\texttt{Texts}})~\footnote{https://cloud.google.com/vision/docs/detecting-safe-search} detects and recognize text within an image, which returns their bounding boxes and transcript texts.
    \item Explicit Content Detection (\textbf{\texttt{SafeSearch}})~\footnote{https://cloud.google.com/vision/docs/detecting-objects} detects explicit content such as adult or violent content within an image, and returns the likelihood.
\end{itemize}

\begin{figure}
\centering
\vspace{0.5cm}
\includegraphics[width=\columnwidth]{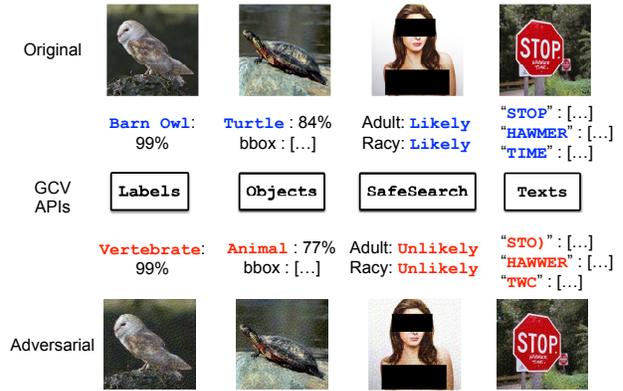}
\caption{Visualization of images chosen from testing set and their corresponding AEs generated by DR. All the AEs are generated on VGG-16 \texttt{conv3.3} layer, with perturbations clipped by $l_\infty\leq16$, and they effectively fool the four GCV APIs as indicated by their outputs.}
\label{fig:demo}
\end{figure}

\textbf{Datasets.} We use ImageNet validation set for testing \texttt{Labels} and \texttt{Objects}, and the NSFW Data Scraper~\cite{nsfw_dataset} and COCO-Text~\cite{icdar} dataset for evaluating against \texttt{SafeSearch} and \texttt{Texts} respectively. We randomly choose 100 images from each dataset for our evaluation, and  Fig.~\ref{fig:demo} shows sample images in our testing set.

\begin{table*}[h]
\centering
\begin{threeparttable}[b]
\begin{tabular}{|c|c|ccccc|}
\hline
\multirow{2}{*}{Model} & \multirow{2}{*}{Attack} & \multicolumn{1}{c}{\textbf{\texttt{Labels}}} & \multicolumn{1}{c}{\textbf{\texttt{Objects}}} & \multicolumn{1}{c}{\textbf{\texttt{SafeSearch}}} & \multicolumn{2}{c|}{
\textbf{\texttt{Texts}}} \\ \cline{3-7} 
 &  & \multicolumn{1}{c}{acc.} & \multicolumn{1}{c}{mAP(IoU=0.5)} & \multicolumn{1}{c}{acc.} & \multicolumn{1}{c}{AP(IoU=0.5)} & \multicolumn{1}{c|}{C.R.W\tnote{2}} \\ \hline \hline
\multicolumn{2}{|c|}{baseline (SOTA)\tnote{1}} & 82.5\% & 73.2 & 100\% & 69.2 & 76.1\% \\ \hline \hline
\multirow{3}{*}{VGG-16} & MI-FGSM & 41\% & 42.6 & 62\% & 38.2 & 15.9\% \\ 
 & DIM & 39\% & 36.5 & 57\% & 29.9 & 16.1\% \\ 
 & DR (\textbf{Ours}) & \textbf{23}\% & \textbf{32.9} & \textbf{35}\% & \textbf{20.9} & \textbf{4.1}\% \\ \hline \hline
\multirow{3}{*}{Resnet-152} & MI-FGSM & 37\% & 41.0 & 61\% & 40.4 & 17.4\% \\ 
 & DIM & 49\% & 46.7 & 60\% & 34.2 & 15.1\% \\
 & DR (\textbf{Ours}) & \textbf{25}\% & \textbf{33.3} & \textbf{31}\% & \textbf{34.6} & \textbf{9.5}\% \\ \hline
\end{tabular}
\begin{tablenotes}
    \item[1] The baseline performance of GCV models cannot be measured due to the mismatch between original labels and labels used by Google. We use the GCV prediction results on original images as ground truth, thus the baseline performance should be 100\% for all accuracy and 100.0 for mAP and AP. Here we provide state-of-the-art performance~\cite{ILSVRC2017, keras, icdar, nsfw_dataset} for reference. 
    \item[2] Correctly recognized words (C.R.W)~\cite{icdar}. 
\end{tablenotes}
\vspace{0.2cm}
\caption{\textbf{The degraded performance of four Google Cloud Vision models, where we attack a single model from the left column.} Our proposed DR attack degrades the accuracy of \textbf{\texttt{Lables}} and \texttt{\textbf{SafeSearch}} to 23\% and 35\%, the mAP of \textbf{\texttt{Objects}} and \textbf{\texttt{Texts}} to 32.9 and 20.9, the word recognition accuracy of \textbf{\texttt{Texts}} to only 4.1\%, which outperform existing attacks. 
}
\end{threeparttable}
\label{tab:result}
\end{table*}

\textbf{Experiment setup.} We choose normally trained VGG-16 and Resnet-152 as our target models, from which the AEs are generated, as Resnet-152 is commonly used by MI-FGSM and DIM for generation~\cite{xie2018improving, dong2018boosting}. As DR attack targets on specific layer, we choose \texttt{conv3.3} for VGG-16 and \texttt{conv3.8.3} for Resnet-152 as per the profiling result in Table~\ref{tab:methodology}. 

\textbf{Attack parameters.} We follow the default settings in~\cite{dong2018boosting} with the momentum decay factor $\mu=1$ when implementing the MI-FGSM attack. For the DIM attack, we set probability $p=0.5$ for the stochastic transformation function $T(x;p)$ as used in~\cite{xie2018improving}, and use the same decay factor $\mu=1$ and total iteration number $N=20$ as in the vanilla MI-FGSM. For our proposed DR attack, we don't rely on FGSM method, and instead we use Adam optimizer ($\beta_1=0.98$, $\beta_2=0.99$) with learning rate $5e^{-2}$ to reduce the dispersion of target feature map. The maximum perturbation of all attacks in the experiment are limited by clipping at $l_\infty=16$, which is still considered less perceptible for human observers~\cite{luo2015foveation}. 

\textbf{Evaluation metrics.} We perform adversarial attacks only on single network and test them on the four black-box GCV models. The effectiveness of attacks are measured by the model performance under attacks. As the labels from original datasets are different from labels used by GCV, we use the prediction results of GCV APIs on the original data as the ground truth, which gives a baseline performance of 100\% accuracy or 100.0 mAP and AP respectively. We also provide state-of-the-art results on each CV tasks as references (Table~\ref{tab:result}). 

Figure~\ref{fig:demo} shows example of each GCV model's output for original and adversarial examples. The performance of \textbf{\texttt{Labels}} and \textbf{\texttt{SafeSearch}} are measured by the accuracy of classifications. More specifically, we use \emph{top\-1} accuracy for \textbf{\texttt{Labels}}, and use the accuracy for detecting our given porn images as \texttt{LIKELY} or \texttt{VERY\_LIKELY} being \texttt{adult} for \textbf{\texttt{SafeSearch}}. 

The performance of \textbf{\texttt{Objects}} is given by the mean average precision (mAP) at $IoU=0.5$. For \textbf{\texttt{Texts}}, we follow the bi-fold evaluation method of ICDAR 2017 Challenge~\cite{icdar}. We measure text localization accuracy using average precision (AP) of bounding boxes at $IoU=0.5$, and evaluate the word recognition accuracy with correctly recognized words (C.R.W) that are case insensitive. 
   
\textbf{Results.}
As shown in Table~\ref{tab:result}, DR outperforms other baseline attacks by degrading the target model performance by a larger margin. For example, the adversarial examples crafted by DR on VGG-16 model brings down the accuracy of  \textbf{\texttt{Labels}} to only 23\%, and \textbf{\texttt{SafeSearch}} to 35\%. Adversarial examples created with the same technique also degrade mAP of \textbf{\texttt{Objects}} to 32.9\% and AP of text localization to 20.9\%, and with barely 4.1\% accuracy in recognizing words. Strong baselines like MI-FGSM and DIM on the other hand, only obtains 38\% and 43\% success rate when attacking \texttt{SafeSearch}, and are less effective compared with DR when attacking all other GCV models. The results demonstrates the better cross-task transferability of dispersion reduction attack. 

When comparing the effectiveness of attacks on different generation models, the results that DR generates adversarial examples that transfer better across these four commercial APIs still hold. The visualization in Fig.~\ref{fig:demo} shows that the perturbed images with $l_\infty\leq16$ well maintain their visual similarities with original images, but fools real-world computer vision systems.
\section{Discussion \& Conclusion}
One intuition behind DR attack is that by minimizing the dispersion of feature maps, we are making images ``featureless'', as few features can be detected, if neuron activations are suppressed by perturbing the input (Fig.~\ref{fig:illustration}). Further, with the observation that low level features bear more similarities across CV models, we hypothesis that DR attack would produce transferable adversarial examples when targeted on intermediate convolution layers. Evaluation on four different CV tasks shows that this enhanced attack greatly degrades model performance, and thus would facilitate evasion attacks against even an ensemble of CV-based detection mechanisms. We hope that our proposed attack can serve as benchmark for evaluating robustness of future defense.
\balance
{\small
\bibliographystyle{ieee_fullname}
\bibliography{egbib}

\begin{thebibliography}{10}\itemsep=-1pt

\bibitem{amazon}
{Amazon Rekognition}.
\newblock \href{https://aws.amazon.com/rekognition/}{Link}.

\bibitem{google_engine}
{Google Cloud Machine Learning Engine}.
\newblock \href{https://cloud.google.com/ml-engine/}{Link}.

\bibitem{google}
{Google Cloud Vision}.
\newblock \href{https://cloud.google.com/vision/}{Link}.

\bibitem{icdar}
{ICDAR2017 Robust reading challenge on COCO-Text}.
\newblock \href{http://rrc.cvc.uab.es/?ch=5&com=evaluation&task=1}{Link}.

\bibitem{ILSVRC2017}
{ImageNet Challenge 2017}.
\newblock \href{http://image-net.org/challenges/LSVRC/2017/results}{Link}.

\bibitem{keras}
{Keras Applications}.
\newblock \href{https://keras.io/applications/}{Link}.

\bibitem{nsfw_dataset}
{NSFW Data Scraper}.
\newblock \href{https://github.com/alexkimxyz/nsfw_data_scraper}{Link}.

\bibitem{brendel2017decision}
Wieland Brendel, Jonas Rauber, and Matthias Bethge.
\newblock Decision-based adversarial attacks: Reliable attacks against
  black-box machine learning models.
\newblock {\em arXiv preprint arXiv:1712.04248}, 2017.

\bibitem{carlini2017towards}
Nicholas Carlini and David Wagner.
\newblock Towards evaluating the robustness of neural networks.
\newblock In {\em 2017 IEEE Symposium on Security and Privacy (SP)}, pages
  39--57. IEEE, 2017.

\bibitem{dong2018boosting}
Yinpeng Dong, Fangzhou Liao, Tianyu Pang, Hang Su, Jun Zhu, Xiaolin Hu, and
  Jianguo Li.
\newblock Boosting adversarial attacks with momentum.
\newblock In {\em Proceedings of the IEEE Conference on Computer Vision and
  Pattern Recognition}, pages 9185--9193, 2018.

\bibitem{goodfellow2014explaining}
Ian~J Goodfellow, Jonathon Shlens, and Christian Szegedy.
\newblock Explaining and harnessing adversarial examples.
\newblock {\em arXiv preprint arXiv:1412.6572}, 2014.

\bibitem{ilyas2018black}
Andrew Ilyas, Logan Engstrom, Anish Athalye, and Jessy Lin.
\newblock Black-box adversarial attacks with limited queries and information.
\newblock {\em arXiv preprint arXiv:1804.08598}, 2018.

\bibitem{kurakin2016adversarial}
Alexey Kurakin, Ian Goodfellow, and Samy Bengio.
\newblock Adversarial machine learning at scale.
\newblock {\em arXiv preprint arXiv:1611.01236}, 2016.

\bibitem{kurakin2018adversarial}
Alexey Kurakin, Ian Goodfellow, Samy Bengio, Yinpeng Dong, Fangzhou Liao, Ming
  Liang, Tianyu Pang, Jun Zhu, Xiaolin Hu, Cihang Xie, et~al.
\newblock Adversarial attacks and defences competition.
\newblock In {\em The NIPS'17 Competition: Building Intelligent Systems}, pages
  195--231. Springer, 2018.

\bibitem{lee2009convolutional}
Honglak Lee, Roger Grosse, Rajesh Ranganath, and Andrew~Y Ng.
\newblock Convolutional deep belief networks for scalable unsupervised learning
  of hierarchical representations.
\newblock In {\em Proceedings of the 26th annual international conference on
  machine learning}, pages 609--616. ACM, 2009.

\bibitem{liu2016delving}
Yanpei Liu, Xinyun Chen, Chang Liu, and Dawn Song.
\newblock Delving into transferable adversarial examples and black-box attacks.
\newblock {\em arXiv preprint arXiv:1611.02770}, 2016.

\bibitem{luo2015foveation}
Yan Luo, Xavier Boix, Gemma Roig, Tomaso Poggio, and Qi Zhao.
\newblock Foveation-based mechanisms alleviate adversarial examples.
\newblock {\em arXiv preprint arXiv:1511.06292}, 2015.

\bibitem{statistics}
Chris~A Mack.
\newblock {\em NIST,SEMATECH e-Handbook of Statistical Methods}.
\newblock 2007.

\bibitem{narodytska2016simple}
Nina Narodytska and Shiva~Prasad Kasiviswanathan.
\newblock Simple black-box adversarial perturbations for deep networks.
\newblock {\em arXiv preprint arXiv:1612.06299}, 2016.

\bibitem{papernot2016transferability}
Nicolas Papernot, Patrick McDaniel, and Ian Goodfellow.
\newblock Transferability in machine learning: from phenomena to black-box
  attacks using adversarial samples.
\newblock {\em arXiv preprint arXiv:1605.07277}, 2016.

\bibitem{papernot2017practical}
Nicolas Papernot, Patrick McDaniel, Ian Goodfellow, Somesh Jha, Z~Berkay Celik,
  and Ananthram Swami.
\newblock Practical black-box attacks against machine learning.
\newblock In {\em Proceedings of the 2017 ACM on Asia conference on computer
  and communications security}, pages 506--519. ACM, 2017.

\bibitem{sharif2016accessorize}
Mahmood Sharif, Sruti Bhagavatula, Lujo Bauer, and Michael~K Reiter.
\newblock Accessorize to a crime: Real and stealthy attacks on state-of-the-art
  face recognition.
\newblock In {\em Proceedings of the 2016 ACM SIGSAC Conference on Computer and
  Communications Security}, pages 1528--1540. ACM, 2016.

\bibitem{su2019one}
Jiawei Su, Danilo~Vasconcellos Vargas, and Kouichi Sakurai.
\newblock One pixel attack for fooling deep neural networks.
\newblock {\em IEEE Transactions on Evolutionary Computation}, 2019.

\bibitem{szegedy2013intriguing}
Christian Szegedy, Wojciech Zaremba, Ilya Sutskever, Joan Bruna, Dumitru Erhan,
  Ian Goodfellow, and Rob Fergus.
\newblock Intriguing properties of neural networks.
\newblock {\em arXiv preprint arXiv:1312.6199}, 2013.

\bibitem{xie2018improving}
Cihang Xie, Zhishuai Zhang, Jianyu Wang, Yuyin Zhou, Zhou Ren, and Alan Yuille.
\newblock Improving transferability of adversarial examples with input
  diversity.
\newblock {\em arXiv preprint arXiv:1803.06978}, 2018.

\bibitem{zhou2018transferable}
Wen Zhou, Xin Hou, Yongjun Chen, Mengyun Tang, Xiangqi Huang, Xiang Gan, and
  Yong Yang.
\newblock Transferable adversarial perturbations.
\newblock In {\em Proceedings of the European Conference on Computer Vision
  (ECCV)}, pages 452--467, 2018.

\end{thebibliography}
}

\end{document}